\documentclass[runningheads]{llncs}
\usepackage{graphicx}
\usepackage{comment}
\usepackage{amsmath,amssymb} % define this before the line numbering.
\usepackage{color}

\usepackage{subcaption}
\usepackage{booktabs}
\usepackage{algorithm}
\usepackage{algpseudocode}
\usepackage{color}
\usepackage{mathtools}

\usepackage{bm}
\usepackage{enumitem}
\usepackage{epsfig}
\usepackage{epstopdf}
\usepackage{amssymb}
\usepackage{pifont}
\usepackage{multirow}

\begin{document}
% \linenumbers
\pagestyle{headings}
\mainmatter
\def\ECCVSubNumber{4367}  % Insert your submission number here

\title{Towards Train-Test Consistency for Semi-supervised Temporal Action Localization} % Replace with your title

%\begin{comment}
\titlerunning{work in progress} 
%\authorrunning{ECCV-20 submission ID \ECCVSubNumber} 
%\author{Anonymous ECCV submission}
%\institute{Paper ID \ECCVSubNumber}
\author{Xudong Lin \quad Zheng Shou \quad Shih-Fu Chang\\}
\institute{Columbia University\\
New York, NY 10027, United States\\
\email{ \{xudong.lin, zs2262, sc250\}@columbia.edu}}
%\end{comment}
%******************

\newcommand{\zs}[1]{\textcolor{blue}{[Zheng: #1]}}

\begin{comment}
\titlerunning{Abbreviated paper title}
\author{Xudong Lin \quad Zheng Shou \quad Shih-Fu Chang\\}
\institute{Columbia University\\
New York, NY 10027, United States\\
\email{ \{xudong.lin, zs2262, sc250\}@columbia.edu}}
\end{comment}
%******************
\maketitle

\begin{abstract}
Recently, Weakly-supervised Temporal Action Localization (WTAL) has been densely studied but there is still a large gap between weakly-supervised models and fully-supervised models. It is practical and intuitive to annotate temporal boundaries of a few examples and utilize them to help WTAL models better detect actions. However, the train-test discrepancy of action localization strategy prevents WTAL models from leveraging semi-supervision for further improvement. At training time, attention or multiple instance learning is used to aggregate predictions of each snippet for video-level classification; at test time, they first obtain action score sequences over time, then truncate segments of scores higher than a fixed threshold, and post-process action segments. The inconsistent strategy makes it hard to explicitly supervise the action localization model with temporal boundary annotations at training time. In this paper, we propose a Train-Test Consistent framework, TTC-Loc. In both training and testing time, our TTC-Loc localizes actions by comparing scores of action classes and predicted threshold, which enables it to be trained with semi-supervision. By fixing the train-test discrepancy, our TTC-Loc significantly outperforms the state-of-the-art performance on THUMOS'14, ActivityNet 1.2 and 1.3 when only video-level labels are provided for training. With full annotations of only one video per class and video-level labels for the other videos, our TTC-Loc further boosts the performance and achieves 33.4\% mAP (IoU threshold 0.5) on THUMOS's 14.

\end{abstract}

\begin{figure}
\centering
\includegraphics[width=1\textwidth]{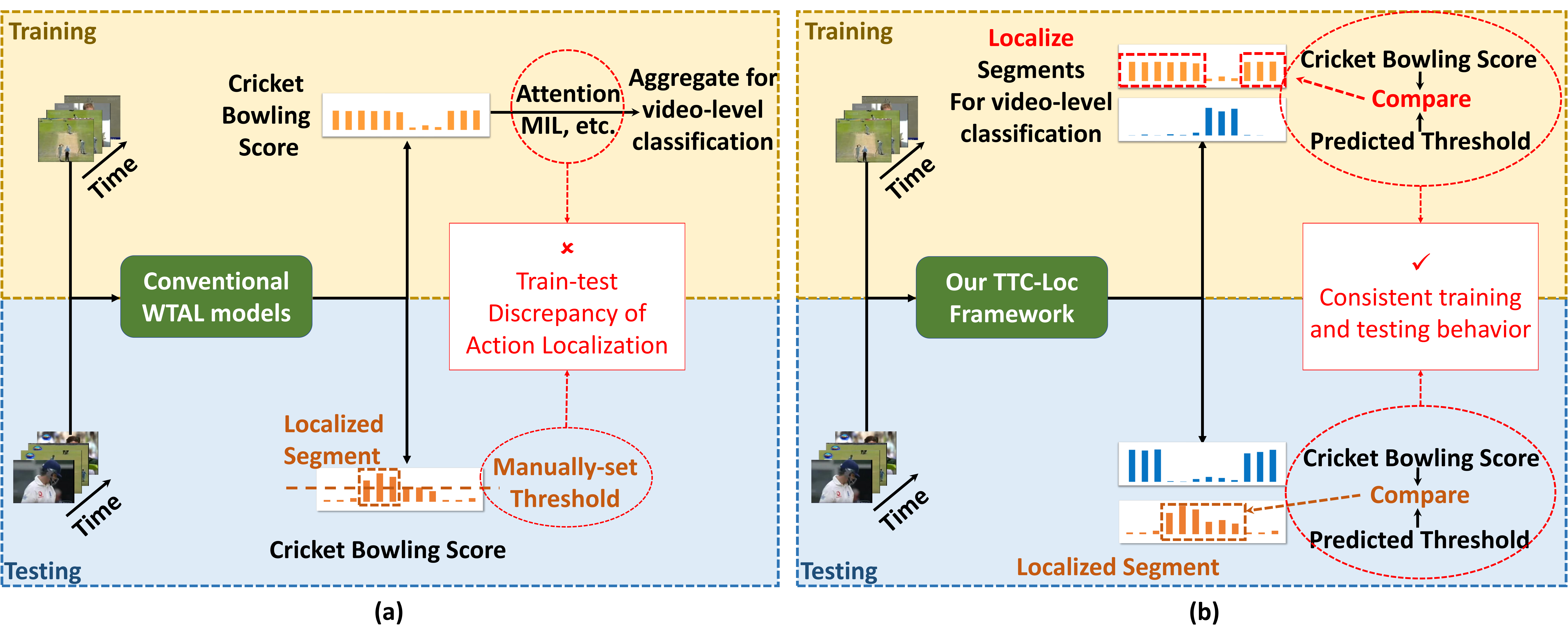}
\caption{(a) Conventional WTAL models are trained to predict per-snippet action classification scores, which are aggregated for video-level classification. At test time, the thresholds for localization are manually set. However, during training, the same threshold is not utilized to localize action segments. (b) We propose to localize action segments with predicted thresholds for both training and testing.
}
\label{fig:idea}
\end{figure}

%%%%%%%%% BODY TEXT
\section{Introduction}
In spite of the impressive progress on fully-supervised temporal action localization \cite{fast_temporal_activity_cvpr16,scnn_shou_wang_chang_cvpr16,iccv17_tap,xu2017r,bmvc17_tad,buch2017sst,liu2016ssd,redmon2016yolo9000,redmon2016you,lin2017single,zhao2017temporal,iccv17_tap,xu2017r,bmvc17_tad}, there is an urgent need to move from the fully-supervised setting to the weakly-supervised setting because annotating the start time and end time for each action segment in all the training videos is too costly and thus is not scalable.
Weakly-supervised Temporal Action Localization (WTAL) is to localize actions and classify them at test time with only the video-level action label annotations during training. However, WTAL models are limited by the lack of temporal boundary annotations and have a large performance gap with fully-supervised models. Motivated by recent success on semi-supervised learning~\cite{papandreou2015weakly,Wei_2018_CVPR,yang2016revisiting,xie2019self,yalniz2019billion,mahajan2018exploring,ji2019learning}, one practical and intuitive solution is annotating the temporal boundaries of a few samples and leveraging them to provide guidance on accurate action detection. In this paper, such a setting, where temporal boundary annotations of a few videos and video-level labels of the rest of the videos are provided during training, is called Semi-supervised Temporal Action Localization (STAL).

One widely-adapted strategy of combining weak and strong supervision is to jointly optimize the model with mixed supervision~\cite{papandreou2015weakly,Wei_2018_CVPR}. For example, in~\cite{papandreou2015weakly}, both pixel-level annotations of a few examples and image-level labels of the rest of images are used to jointly optimize the image segmentation model. However, current WTAL models cannot be easily adapted for Semi-supervised training because of the train-test discrepancy of action localization. As Fig. \ref{fig:idea}(a) shows, the test time action detection strategy adopted by conventional WTAL work \cite{wang2017untrimmednets,singh2017hide,nguyen2018weakly,paul2018w,Liu_2019_CVPR,narayan20193cnet} is first obtaining action score sequences over time, then truncating segments of scores higher than a fixed threshold at every snippet and applying post-processing techniques to obtain action segments. At training time, the model, nevertheless, does not detect action segments in this way but just aggregate scores from snippets for video-level classification via attention or multiple instance learning (MIL), which is inconsistent with the model's test time behavior. Therefore, temporal boundary annotations cannot be explicitly used to supervise the action localization model at training time because of inconsistent strategies for training and testing. Moreover, a manually-set threshold is not flexible for localizing action boundaries because a low threshold may lead false positive instances while a high threshold may over-segment one instance into several segments.

To fix the train-test discrepancy in existing WTAL methods, as shown in Fig.~\ref{fig:idea}(b), we propose a Train-Test Consistent framework, TTC-Loc, which predicts adaptive thresholds to detect action segments in the same manner at both training and test time. When without temporal boundary annotations, our TTC-Loc is able to predict accurate thresholds compared to manually set one because of the consistent train-test action localization strategy. With localized action segments at training time, temporal boundary annotations of a few examples can directly supervise our TTC-Loc and help it to better detect actions.

Our contributions lie in three-fold.
\begin{itemize}
    \item To the best of our knowledge, we make the first attempt to fix the train-test discrepancy in WTAL methods by abandoning the manually set threshold and predicting an adaptive threshold at both training and testing time.
    \item We propose Semi-supervised Temporal Action Localization, which boosts the performance of WTAL models by utilizing both video-level labels and temporal boundary annotations of a few examples for training.
    \item Our method significantly outperforms the state-of-the-art WTAL methods on three challenging benchmark datasets THUMOS'14, ActivityNet 1.2 and ActivityNet 1.3; under Semi-supervised setting, our TTC-Loc further improves the performance and achieves 33.4\% mAP (IoU threshold 0.5) on THUMOS 14 with only one video of full annotation in each class.
\end{itemize}

%-------------------------------------------------------------------------

\section{Related Work}

%\subsection{Video Action Recognition}
% It serves as an effective backbone network in various video analysis tasks, e.g., recognition \cite{Kinetics,shou2019dmc}, segmentation \cite{nilsson2018semantic}, localization \cite{chao2018rethinking}, and etc.

\subsection{Temporal Action Localization}
In the past decade, many backbone deep neural networks have been proposed for image analysis \cite{spp,He_2016_CVPR,huang2017densely} and video analysis or sequence modeling \cite{3dcnn,ji,tran2017convnet,lrcn2014,Simonyan14b,Kinetics,TSN,lin2019contextgated}. These networks have achieved significant improvement on many tasks compared to the conventional methods. For instance, I3D \cite{Kinetics} borrows the idea of Inception structure \cite{szegedy2015going,szegedy2016rethinking} and extends it to 3D to perform spatial-temporal modeling. There are several video datasets for Temporal Action Localization (TAL) such as Charades \cite{Charades1,Charades2}, ActivityNet \cite{caba2015activitynet}, THUMOS \cite{THUMOS14,THUMOS15}. Based on these backbone CNNs and datasets, impressive work has been done on fully-supervised TAL \cite{scnn_shou_wang_chang_cvpr16,iccv17_tap,xu2017r,bmvc17_tad,buch2017sst,lin2017single,zhao2017temporal,iccv17_tap,xu2017r,bmvc17_tad}. For instance, inspired by single-shot object detection method \cite{liu2016ssd,redmon2016yolo9000}, in \cite{iccv17_tap}, direct boundary prediction via anchor generation and boundary regression has been adapted from object detection in order to detect more accurate boundaries.

\subsection{Weakly-supervised Temporal Action Localization}

Obtaining the temporal annotations for full supervision is still the bottleneck if we go to a larger scale. Data with only video level annotations is much cheaper. Therefore, it is practical and interesting to explore WTAL models with only video-level annotations \cite{sssn_mm15,wang2017untrimmednets,singh2017hide,nguyen2018weakly,paul2018w,xu2018segregated,Zhong_2018,Liu_2019_CVPR,narayan20193cnet}. Different types of weak supervision have also been explored. For instance, in \cite{huang2016connectionist,richard2017weakly} the order of actions is taken as the extra supervision.
%Mettes \textit{et al.} \cite{mettes2016spot} proposed to train a spatial-temporal action detector with point-level supervision. 
%In this paper, we are focused on the setting, where only video action labels are provided.

UntrimmedNets~\cite{wang2017untrimmednets} consists of a classification module and a selection module. It is only constrained by the video-level classification loss, which makes the detection model only capable of detecting the most discriminative part of an action \cite{singh2017hide}. STPN \cite{nguyen2018weakly} further constrains the selection module, i.e., the attention module, with sparsity assumption. However, they fail to model the relationship between different classes or videos, which leaves room for improvement.
%AutoLoc \cite{shou2018autoloc} emphasizes the weakness of using a manually set threshold in aforementioned methods and learns a model which directly predicts the boundary from class activation sequence. 

ST-GradCAM \cite{xu2018segregated} borrows the idea from natural language processing to tackle the attention overlapping issue, and utilizes an enhanced Recurrent Neural Network to model the temporal dependency. WTALC \cite{paul2018w} and ST-GradCAM \cite{xu2018segregated} consider inter-class or inter-video relationships. 
However, they fail to directly model the existence of backgrounds in videos and thus are unable to utilize the informative prior knowledge between actions and backgrounds. Most of current WTAL methods employ a manually-set threshold to perform localization. Shou \textit{et al.} \cite{shou2018autoloc} proposed to directly regress the boundary of action segment, which is motivated by aforementioned boundary regression methods in fully-supervised manner. However, it is unclear how to train this framework in an end-to-end manner with video-level labels. In \cite{Zhong_2018,su2018cascaded}, erasing operation is introduced to generate fine boundaries. However, to the best of our knowledge, learning to predict threshold hasn't been explored in WTAL. 

\subsection{Semi-supervised Learning}
Semi-supervised learning~\cite{papandreou2015weakly,Wei_2018_CVPR,yang2016revisiting,xie2019self,yalniz2019billion,mahajan2018exploring,ji2019learning} has been proven to be very effective in improving models' generalization ability. In \cite{ji2019learning}, a semi-supervised temporal proposal generation model is trained by both proposal labels and teacher-student consistency towards temporal perturbations. However, in their setting, the dataset consists of videos with proposal annotations and unlabeled videos, which differs from our semi-supervised training setting. Among different settings of semi-supervised learning, \cite{papandreou2015weakly,Wei_2018_CVPR} are most relevant to ours. We all aim to improve weakly-supervised models' performance with a few fully annotated samples. However, their task \cite{papandreou2015weakly,Wei_2018_CVPR} is image segmentation. To the best of our knowledge, we are the first to formulate semi-supervised learning for temporal action localization in this way.

\section{Proposed Approach}

%Previous WTAL methods failed to explicitly model the informative relationships between backgrounds and ground truth actions.
%Our approach is based on a simple observation: one frame is supposed to be assigned to either backgrounds or ground truth action classes.
%Based on this observation, we propose aforementioned three constraints on weakly supervised action localization models. In the following of 
In this section, we will first go through the main components of our model, then present the training strategy of our model with weak/semi-weak supervision, and finally describe how to detect actions at the test time.
%\begin{itemize}
%	\item Time stamps which have larger activation for a class than for background are supposed to be allocated for video-level classification;
%	\item Activated time of background and groundtruth action classes is supposed to be complementary;
%	\item The similarity between two videos' features when background is activated is supposed to be larger than the similarity between the features of one video when one action class is activated and the features of another video when background is activated.
%\end{itemize} 

\begin{figure*}[tb]
\centering
\includegraphics[width=1\textwidth]{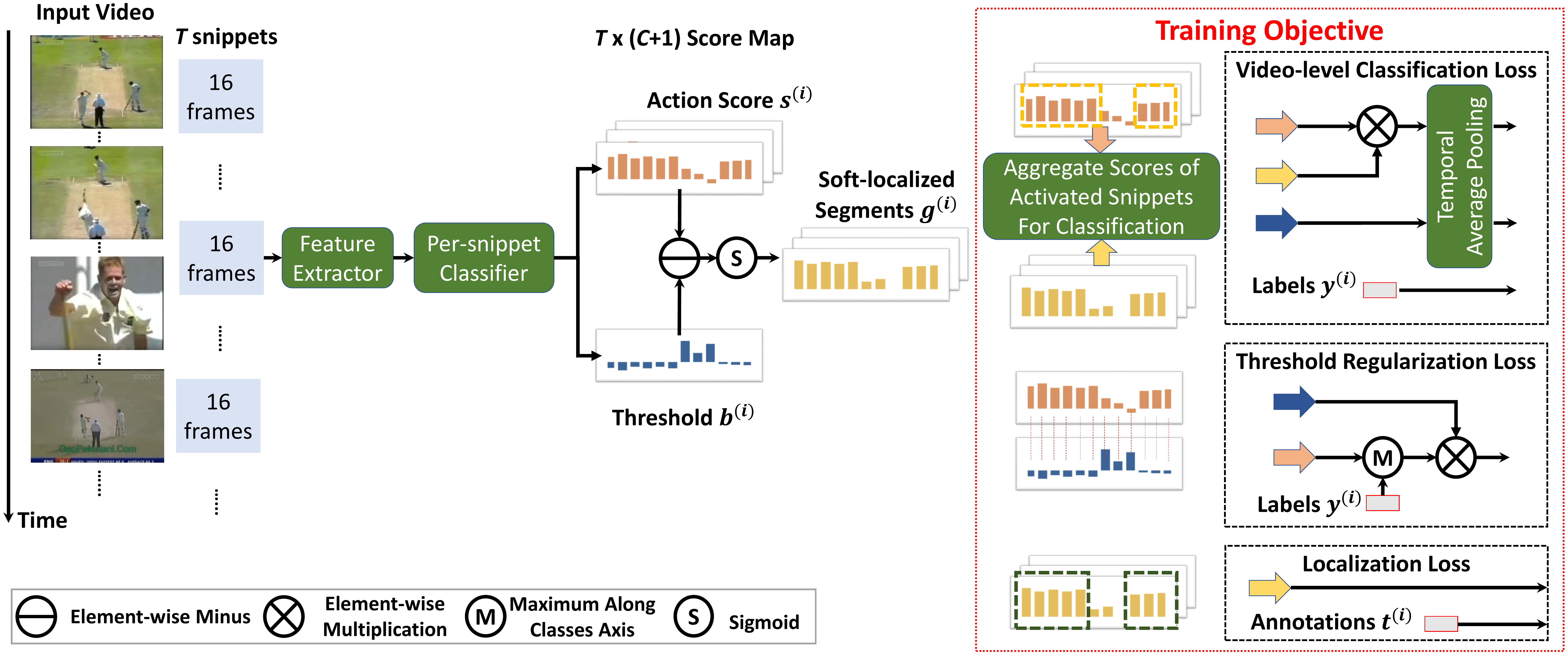}
\caption{The pipeline of the proposed TTC-Loc framework. We first use a feature extractor to extract features for snippets; then we build a per-snippet classifier to produce scores $s^{(i)}$ for each snippet. To train the network, we first localize action segments with $g^{(i)}$ and allocate action scores as the video-level predictions; for weakly-supervised setting, Video-level Classification Loss and Threshold Regularization Loss are employed to enforce the model to classify videos correctly and produce reasonable thresholds; under semi-supervised setting, Localization Loss is also used to jointly optimize the model. (Best viewed in color)
} 
\label{fig:framework}
\end{figure*}

\subsection{Pipeline}
%\textbf{Feature Extraction.} 
Fig. \ref{fig:framework} shows the whole pipeline of our proposed TTC-Loc framework. Following \cite{nguyen2018weakly,paul2018w}, we take I3D network or UntrimmedNets to extract features for non-overlapping snippets. Each snippet consists of 16 frames. We extract two 1,024-dimension feature vectors respectively from RGB frames and corresponding optical flow. Then we concatenate them into a 2,048-dimension feature vector for each snippet. Formally, given $\mathbf{X}={({\mathbf{x}^{(1)}, \cdots, \mathbf{x}^{(n)}})}$ as inputs, we use a feature extractor $E$ to extract features $E(\mathbf{x}^{(i)})$ for each snippet. $E(\mathbf{x}^{(i)})$ is a $T \times D$ matrix, and $D$ is the dimension of the feature vectors (here $D=2048$) and $T$ is the number of snippets. We build a three-layer classification network $F$ to predict $(C+1)$-dimension score vectors. $C$ is the number of action classes and the additional one is for threshold. 

%Previous work either classifies each snippet into $C$ actions~\cite{wang2017untrimmednets,nguyen2018weakly,paul2018w} or collects statics videos for background class~\cite{Liu_2019_CVPR}. However, our model directly takes the predicted background score as adaptive threshold at both training and test time.

%\zheng{I feel can remove this paragraph. this is not clear and also redundant to 3.2.1}
%Since we directly model the background, we can utilize it to realize gating mechanism. By comparing the scores of action and background, we can build a temporal-class-wise gate which performs selection along the temporal axis. The combination of Gate mechanism and temporal average pooling is Gated Temporal Average Pooling, which helps us to select relevant snippets.

%\zheng{I feel can remove this paragraph.}
%At the training time, our model only has video-level action labels as supervision and it is able to detect action segments because of our three constraining loss terms.
%At the testing time, we first use the network to produce video level classification score and localize the actions with top scores.

%\zheng{now the paper uses frame, snippet, time stamp. would be good to make all consistent somehow.}

\subsection{Training}
\subsubsection{Video-level Classification Loss.}
\label{ce}
In order to aggregate per-snippet scores for video classification, previous methods usually rely on the predicted attention \cite{wang2017untrimmednets,nguyen2018weakly} or select top $k$ action scores \cite{paul2018w}, rather than aggregating from localized action segments. With the predicted threshold, our TTC-Loc first localizes action segments for each class by comparing the action score and predicted threshold at each snippet. However, using hard assignment for each snippet is difficult for optimizing the network. To ease the training process, inspired by previous works on deep hashing/binary encoding~\cite{erin2015deep,duan2018graphbit,karaman2019unsupervised}, we relax the localization process with a Sigmoid function. Detailed study on other relaxing functions are provided in Section~\ref{gate_form}. Formally, for each video sample $\mathbf{x}^{(i)}$, we first build a $T \times C$ gate as the soft action localization results,
\begin{eqnarray}
\label{eqn:gate}
g^{(i)}_{(c)} = Sigmoid(s^{(i)}_{(c)} - b^{(i)}), \quad c=1,...,C.
\end{eqnarray}
where $s^{(i)}$ is the $T \times C$ score map consisting of scores
of each action class at each snippet, $b^{(i)}$ is the threshold vector and $Sigmoid(x) \coloneqq 1/(1+e ^{-x})$. With this temporal-class-wise gate, we perform average pooling to get the fused score for video-level action classification,
\begin{eqnarray}
\hat{s}^{(i)}_{(c)} = \frac{1}{\sum _{t=1}^Tg^{(i)}_{(tc)}}\sum _{t=1}^T g^{(i)}_{(tc)}s^{(i)}_{(tc)}, \quad c=1,...,C.
\end{eqnarray}
$\hat{s}^{(i)}_{(c)} $ is the temporal average score for class $c$. 

Based on the observation that a snippet can be either classified as actions or background (no action happens), background score and the predicted threshold are equivalent in our formulation. Therefore, we also apply temporal average pooling to predicted threshold and regard it as a background score,
\begin{eqnarray}
\hat{b}^{(i)} = \frac{1}{T}\sum _{t=1}^T b^{(i)}_{(t)}.
\end{eqnarray}

After temporal pooling, we utilize Softmax function over action classes and background to obtain the final predicted probability,
\begin{align}
\label{eqn:proba}
p_{(c)}^{(i)} =
\begin{cases}
 \frac{\exp{(\hat{s}^{(i)}_{(j)})}}{\exp(\hat{b}^{(i)})+\sum_{j=1}^{C}\exp{(\hat{s}^{(i)}_{(j)})}} \quad c=1,...,C \\
\frac{\exp{(\hat{b}^{(i)})}}{\exp(\hat{b}^{(i)})+\sum_{j=1}^{C}\exp{(\hat{s}^{(i)}_{(j)})}} \quad c=C+1.
\end{cases} 
\end{align}

Then We apply Video-level Classification Loss to enable the model to classify the video correctly. Formally, for a batch consisting of $B$ samples,
\begin{eqnarray}
L_{clas} =- \frac{1}{B}\sum_{i=1}^{B}\sum_{j=1}^C (y^{(i)}_{(j)} \log p^{(i)}_{(j)} + w_b \log p^{(i)}_{(C+1)}),
\end{eqnarray}
where $w_b$ is the weight for the entropy of background, and $y^{(i)}$ is a normalized multi-hot label vector. Note that for untrimmed videos, we can confidently assume that most of the videos have background segments (where no action happens and predicted threshold is supposed to be larger than action scores of any classes. But average amount of the video samples from one action class is likely to be much smaller than the size of the whole dataset. Due to the extreme imbalance between background and action classes, we need to give background smaller weight to avoid the model only detects background. Without more statistics of the dataset, we heuristically set $w_b=1/C$.

\subsubsection{Threshold Regularization Loss.}
\label{cs}
The Video-level Classification Loss only constrains the model to recognize actions and backgrounds from each video, but there is no constraint on the predicted threshold. An intuitive observation is that when the model can confidently assign a snippet to either background or foreground for the ground-truth action class, the model is more likely to be robust when there are noise and variance at inference time. When the predicted threshold and action scores are contrary with a large margin at each snippet, the generated proposal will have high confidence. To formulate this constraint, we first obtain the union of activated areas of ground truth classes by selecting the maximum value in the score map along class axis at each snippet,
\begin{eqnarray}
\tilde{s}^{(i)}_{(t)}=\max_{y_c>0} s^{(i)}_{(tc)}.
\end{eqnarray}

By forcing product of action score and background score at each snippet to be negative, we can get
\begin{eqnarray}
L_{reg} = \frac{1}{B}\sum_{i=1}^{B} \sum_{t=1}^{T} \frac{\max(\tilde{s}^{(i)}_{(t)} b^{(i)}_{(t)} + 1, 0)}{||\tilde{s}^{(i)}||_2  ||b^{(i)}||_2} ,
\end{eqnarray}
where $||*||_2$ denotes the L-2 norm. This formulation is similar to cosine similarity, but it is more suitable for our case. The cosine similarity has a minimum value when every element of one normalized score vector is the additive inverse of the other one, which brings extra constraint. In our case, as long as they have different signs and a large margin, our constraint has been satisfied. Detailed study on loss forms are provided in Section~\ref{gate_form}.

The Threshold Regularization Loss is crucial for training a better weakly-supervised temporal action localization model. $L_{clas}$ only makes the model suitable for video-level classification. The model may learn to produce proposals with blurred boundaries. $L_{reg}$ guarantees a large margin between action scores and background scores and helps to generate more confident action proposals, which is complementary with $L_{clas}$. 

\subsubsection{Localization Loss.}
For samples with temporal boundary annotations, we can explicitly supervise their soft localization results $g^{(i)}$. The temporal boundary annotations are transformed in to $T \times C$ binary matrices $a^{(i)}$, where 1 represents that corresponding snippet is within an action segment of a certain class. Then we can formulate the Localization loss as follows,
\begin{eqnarray}
L_{loc} = \frac{1}{|S|TC}\sum_{i \in S} \sum_{t=1}^{T} \sum_{t=1}^{C} |g^{(i)}_{(tc)}- a^{(i)}_{(tc)}| ,
\end{eqnarray}
where $S$ is the set of samples with full annotations in this batch.

\subsubsection{Full Objective and Discussion}
With the aforementioned three losses constraining the model from different aspects, we jointly optimize them together with the balancing weights $\lambda$ and $\eta$,
\begin{eqnarray}
\label{eqn:total}
L=\lambda L_{clas}+(1-\lambda)L_{reg} + \eta L_{loc}.
\end{eqnarray}

When only video-level labels are available, the first two terms constrain our proposed TTC-Loc to correctly classify videos and to predict reasonable thresholds; when temporal boundary annotations of a few examples are also provided, the Localization Loss explicitly utilizes the supervision to help our TTC-Loc better detection actions.

We are aware of previous methods~\cite{nguyen2019weakly,Liu_2019_CVPR} that also perform $(C+1)$-way classification including background. However, the train-test discrepancy of action localization still exists in these methods, which is the fundamental difference between our proposed TTC-Loc and aforementioned methods. Consequently, our proposed TTC-Loc explicitly enjoys the benefit of semi-supervised training.
%From the above analysis, the Video-level Classification Loss $L_{clas}$ for classification and the Snippet-level Foreground-Background Score-Contrary Loss $L_{cont}$ for generated proposals are crucial and complementary for the network to be able to correctly detect actions in videos. Based on the assumption and formulation, the predicted background scores can serve as a promising threshold for localization in both training and testing, while previous methods \cite{wang2017untrimmednets,nguyen2018weakly,paul2018w} usually use a manually set threshold for all time stamps. To the best of our knowledge, this is the first work that bridges the gap between training and testing in conventional WTAL methods. 

%: e.g. when the model is over-fitting and for one class, its score is falsely always larger than background score at every time snippet, then the feature representation of the unactivated area will share lower similarity with features from unactivated area of another video, which may be penalized the Metric Learning Loss.

\subsection{Inference}
At inference time, we first binarize the soft localization gate $g^{(i)}$ and then connected non-zero elements are already action segment candidates; then we use the confidence scores obtained in Equation \ref{eqn:proba} to classify the video; we select action classes with confidence scores larger than the average confidence score over $C$ action classes and only keep segment candidates from selected classes. The kept segments are detected segments and then we can obtain corresponding start time, end time and score.

\section{Experiments}
To stress the effectiveness of our TTC-Loc method, we conduct experiments on three challenging datasets: THUMOS'14 \cite{THUMOS14}, ActivityNet 1.2 \cite{caba2015activitynet} and ActivityNet 1.3 \cite{caba2015activitynet}. We will first introduce the datasets and evaluation metric, then provide implementation details, comparisons with state-of-the-art methods, and detailed analysis.
\subsection{Datasets and Evaluation Metric}
\subsubsection{THUMOS'14} 
We follow \cite{wang2017untrimmednets,shou2018autoloc,paul2018w} to conduct experiments on the temporal action localization task of THUMOS'14. Its validation set consists of 200 untrimmed videos, each of which contains at least one action. The test set contains 213 videos. The validation set and test set contain the same 20 classes of actions. According to \cite{wang2017untrimmednets,shou2018autoloc,paul2018w}, we train the model on the validation set and test it on the test set.

\subsubsection{ActivityNet 1.2} 
We follow \cite{wang2017untrimmednets,shou2018autoloc,paul2018w} to use ActivityNet version 1.2 \cite{caba2015activitynet} for comparisons. It contains 100 activity classes. The training set consists of 4,819 videos and the validation set has 2,383 videos. According to \cite{wang2017untrimmednets,shou2018autoloc,paul2018w}, we train the model on the training set and test it on the validation set.

\subsubsection{ActivityNet 1.3} 
ActivityNet version 1.3 \cite{caba2015activitynet} contains 200 activity classes. The training set consists of 10,024 videos and the validation set has 4,926 videos. According to \cite{nguyen2019weakly,xu2018segregated}, we train the model on the training set and test it on the validation set.

\subsubsection{Evaluation Metric}
Following previous works on temporal action localization \cite{wang2017untrimmednets,shou2018autoloc,paul2018w,nguyen2018weakly,narayan20193cnet}, we evaluate our models with mean Average Precision (mAP). Each predicted segment is regarded as correct only when the predicted class is correct and its temporal overlap IoU with ground truth segment is larger than certain threshold. Duplicate detection for the same ground truth segment is not included.

\subsection{Implementation Details}
We implement our TTC-Loc framework with PyTorch \cite{pytorch} and conduct experiments on one NVIDIA GeForce GTX TITAN X GPU. We utilize TV-L1 \cite{zach2007duality} algorithm to extract optical flow. Due to the limitation of GPU memory, we do not fine-tune feature extractor in training. The output dimension of the first fully connected layer is still 2,048. We follow~\cite{Liu_2019_CVPR} to add a temporal-only convolution with a temporal kernel size of 3 and an output dimension of 2048. To ease the training process, we add the temporal convolution in a residual connection manner~\cite{He_2016_CVPR}. The fully connected layer and the convolution layer come with ReLU \cite{krizhevsky2012imagenet} activation function. Before classification, we use Dropout \cite{srivastava2014dropout} with a dropout rate of 0.7 in all the experiments. The output layer is a fully connected layer with linear activation. 

We sample the first k samples in each class for semi-supervision. We also try randomly selecting samples in each class and do not observe significant differences between different subsets of samples used for semi-supervised learning. We randomly sample from the whole dataset to construct mini-batches. We optimize the loss function in Equation \ref{eqn:total} using Adam \cite{kingma2014adam} with a batch size of 10. The weight $w_b$ for background is set to be 0.05 for THUMOS'14, 0.01 for ActivityNet 1.2 and 0.005 for ActivityNet 1.3, according to the number of classes in them. For the learning rate in all the experiments, we start with $10^{-4}$ and do not manually decrease it later. Following \cite{paul2018w}, we set the maximum length of video snippets to be 320 and 750 for THUMOS'14 and ActivityNet 1.2/1.3 respectively; when the video is longer than the maximum length, we randomly extract a clip of the maximum length from it. The balancing factor $\lambda$ is set to be 0.2 and 0.4 on THUMOS'14 and ActivityNet 1.2, respectively. $\eta$ is set to be 3 for all experiments. The detailed experiments on the effects of $\lambda$ and $\eta$ are provided in the supplementary material.

\begin{table*}[!t]
\caption{Comparisons with the state-of-the-art methods on temporal localization mAP (\%) under different IoU thresholds on THUMOS'14 test set. Weak: trained with the video-level labels only. Full: temporal boundaries of action segment are used for training. Semi (k) indicates that k samples per class have full annotations. Untrim indicates UntrimmedNets features and I3D indicates I3D features. Average is the average mAP from IoU 0.3 to 0.7.}
	\label{table:res_th}
	\begin{center}
	\tiny
	
	\begin{tabular}{|c|c|cccccc|}
	\hline
	Setting & IoU threshold $\rightarrow$     & 0.3           & 0.4           & 0.5           & 0.6           & 0.7   & Average       \\ \hline
	%Full        & Karaman \textit{et al.} \cite{th3} &  0.5 &  0.3 &  0.2 &  0.2 &  0.1 & 0.3  \\
	%Full        & Wang \textit{et al.} \cite{th2} &  14.6 &  12.1 &  8.5 & 4.7 &  1.5  & 8.3 \\
	%Full        & Heilbron \textit{et al.} \cite{fast_temporal_activity_cvpr16} &  - &  - &  13.5 &  - &  -   \\
	%Full        & Escorcia \textit{et al.} \cite{victor_eccv16} &  - &  - &  13.9 &  - &  -  \\
	%Full        & Oneata \textit{et al.} \cite{th1}  &  28.8 &  21.8 &  15.0  &  8.5 &  3.2 \\
	%Full        & Richard and Gall \cite{Richard_2016_CVPR} &  30.0 &  23.2 &  15.2  &  - &  -  \\
	%Full        & Yeung \textit{et al.} \cite{stanford_cvpr16} &  36.0 &  26.4  &  17.1 &  - &  -  \\
	%Full        & Yuan \textit{et al.} \cite{yuan_cvpr16} &  33.6 &  26.1 &  18.8  &  - & -   \\
	%Full        & Yuan \textit{et al.}  \cite{yuan2017temporal} &  36.5 &  27.8 & 17.8 &  - &  - &  - \\ 
	Full        & S-CNN  \cite{scnn_shou_wang_chang_cvpr16} &  36.3 &  28.7 & 19.0 &  10.3 &  5.3 & 19.9 \\ 
	%Full        & SST \cite{buch2017sst}          & 37.8          & -          & 23.0          & -          & -     & -     \\
	Full        & CDC \cite{cdc_zheng_cvpr17}          & 40.1          & 29.4          & 23.3          & 13.1          & 7.9      & 22.8    \\
	%Full        & Dai \textit{et al.} \cite{dai2017temporal}          & -          & 33.3          & 25.6          & 15.9       & 9.0   & -\\
	%Full        & SSAD \cite{lin2017single}          & 43.0          & 35.0          & 24.6          & -       & -  &  -\\
	%Full        & TURN TAP \cite{iccv17_tap}          & 44.1          & 34.9          & 25.6          & -       & - &  -\\
	%Full        & R-C3D \cite{xu2017r}          & 44.7          & 35.6          & 28.9          & -       & - \\
	%Full        & SS-TAD \cite{sstad_buch_bmvc17}          & 45.7          & -          & 29.2          & -       & 9.6 &  -\\
	%Full        & Gao \textit{et al.} \cite{bmvc17_tad}          & 50.1          & 41.3          & 31.0          & 19.1       & 9.9 & 30.3 \\
	Full        & SSN    \cite{zhao2017temporal,shou2018autoloc}      & 51.9          & 41.0          & 29.8          & 19.6          & 10.7  & 30.6       \\ 
	Full        & TAL-Net    \cite{chao2018rethinking}      & \textbf{53.2}          & \textbf{48.5}          & \textbf{42.8}          & \textbf{33.8}          & \textbf{20.8}  & \textbf{41.3}       \\ 
	\hline
	%Weak        & Sun \textit{et al.} \cite{sssn_mm15} & 8.5          & 5.2          & 4.4           & -             & -            \\
	Weak        & Hide-and-Seek \cite{singh2017hide} & 19.5          & 12.7          & 6.8           & -             & -         &  -   \\
	Weak        & UntrimmedNets \cite{wang2017untrimmednets} & 28.2          & 21.1          & 13.7          & -             & -    &  -        \\
	\hline
	%Weak, Untrim        & STPN  & 45.3 & 38.8 & 31.1 & 23.5 & 16.2 & 9.8 &  5.1            \\
	%Weak, Untrim        & AutoLoc \cite{shou2018autoloc}       & 35.8 & 29.0 & 21.2 & 13.4 & 5.8 &  21.0 \\
	%Weak, Untrim       & WTALC  & 49.0   & 42.8 &  32.0  &  26.0 & 18.8  & - &  6.2           \\
	%Weak, Untrim       & TTC-Loc (Ours)   & \textbf{53.6} & \textbf{48.6}  & \textbf{38.8}  & \textrbf{30.6} & \textbf{21.8} & \textbf{13.8} & \textbf{7.5}    \\
	Weak, Untrim        & STPN \cite{nguyen2018weakly}  & 31.1 & 23.5 & 16.2 & 9.8 &  5.1     & 17.1       \\
	Weak, Untrim       & AutoLoc \cite{shou2018autoloc}       & 35.8 & 29.0 & 21.2 & 13.4 & 5.8 &  21.0 
	\\
	Weak, Untrim       & WTALC \cite{paul2018w}  &  32.0  &  26.0 & 18.8  & - &  6.2    & -       \\
	Weak, Untrim       & CleanNet \cite{Liu_2019_ICCV} & 36.3 &  30.7 &  22.9 &  13.8 &   5.3 &21.8    \\
	Semi (0), Untrim       & \textbf{TTC-Loc (Ours)}    & \textbf{39.9}  & \textbf{31.5} & \textbf{22.6} & \textbf{14.2} & \textbf{7.9}  & \textbf{23.2}  \\
	\hline
	Weak + Static Clips, Untrim & Liu \textit{et al.}~\cite{Liu_2019_CVPR} & 37.5          & 29.1          & 19.9          & 12.3             & 6.0      &  21.0     \\
	Semi (1), Untrim       & \textbf{TTC-Loc (Ours)}    & \textbf{43.3}  & \textbf{34.4} & \textbf{26.2} & \textbf{17.0} & \textbf{9.4}  & \textbf{26.1}  \\
	Semi (2), Untrim       & \textbf{TTC-Loc (Ours)}    & \textbf{43.3}  & \textbf{35.2} & \textbf{26.9} & \textbf{17.8} & \textbf{9.8}  & \textbf{26.6}  \\
	Semi (3), Untrim       & \textbf{TTC-Loc (Ours)}    & \textbf{43.7}  & \textbf{36.4} & \textbf{27.8} & \textbf{18.0} & \textbf{10.7}  & \textbf{27.3}  \\
	%Semi (4), Untrim       & \textbf{TTC-Loc (Ours)}    & \textbf{44.1}  & \textbf{36.5} & \textbf{28.5} & \textbf{19.0} & \textbf{10.7}  & \textbf{27.8}  \\
	%Full, Untrim       & \textbf{TTC-Loc (Ours)}    & \textbf{44.7}  & \textbf{36.8} & \textbf{28.9} & \textbf{19.8} & \textbf{11.4}  & \textbf{28.4}  \\
	\hline
	Weak, I3D        & STPN \cite{nguyen2018weakly} & 35.5          & 25.8          & 16.9          & 9.9             & 4.3        &  18.5    \\
	
	Weak, I3D        & WTALC \cite{paul2018w} & 40.1          & 31.1          & 22.8          & 14.8             & 7.6    &  23.3        \\

	Weak, I3D        & ST-GradCAM \cite{xu2018segregated} & \textbf{48.7}          & 34.7          & 23.0          & 11.7             & 6.2      &  24.9     \\
	Weak, I3D        & TSM \cite{yu2019temporal} & 39.5         & -         & 24.5         & -             & 7.1    &  -        \\
	
	Weak, I3D        & Nguyen \textit{et al.} \cite{nguyen2019weaklysupervised} & 46.6         & 37.5          & 26.8          & 17.6             & 9.0    &  27.5       \\
	Semi (0), I3D        & \textbf{TTC-Loc (Ours)}     & 46.7  & \textbf{37.6} & \textbf{28.9} & \textbf{17.7} & \textbf{10.0}   &  \textbf{28.2} \\
	\hline
	Weak + Static Clips, I3D & Liu \textit{et al.}~\cite{Liu_2019_CVPR} & 41.2          & 32.1          & 23.1          & 15.0             & 7.0      &  23.7     \\
	Weak + MV, I3D        & Nguyen \textit{et al.} \cite{nguyen2019weaklysupervised} & 49.1        & 38.4         & 27.5         & 17.3             & 8.6    &  28.2      \\
	Weak + Count Labels, I3D        & 3C-Net \cite{narayan20193cnet} & 44.2         & 34.1          & 26.6          & -             & 8.1    &  -        \\
	Semi (1), I3D       & \textbf{TTC-Loc (Ours)}    & \textbf{50.9}  & \textbf{41.9} & \textbf{33.4} & \textbf{21.7} & \textbf{12.1}  & \textbf{32.0}  \\
	Semi (2), I3D       & \textbf{TTC-Loc (Ours)}    & \textbf{51.3}  & \textbf{41.9} & \textbf{33.7} & \textbf{22.9} & \textbf{13.7}  & \textbf{32.7}  \\
	Semi (3), I3D       & \textbf{TTC-Loc (Ours)}    & \textbf{52.8}  & \textbf{44.4} & \textbf{35.9} & \textbf{24.7} & \textbf{13.8}  & \textbf{34.3}  \\
	%Full, I3D       & \textbf{TTC-Loc (Ours)}    & \textbf{55.8}  & \textbf{48.0} & \textbf{41.0} & \textbf{27.7} & \textbf{15.5}  & \textbf{37.6}  \\
	\hline

	\end{tabular}
	
	\end{center}
	\vspace{-0.4cm}
\end{table*}

\begin{table*}[t]
    \caption{Comparisons with the state-of-the-art methods on temporal localization mAP (\%) under different IoU thresholds on ActivityNet 1.2 and 1.3 validation set. Weak: trained with the video-level labels only. Full: temporal boundaries of action segment are used for training. Semi (k) indicates that k samples per class have full annotations. Average is the average mAP from IoU 0.5 to 0.95.}
	\label{table:res_an}
    
	\begin{center}
	\tiny	
		
		\begin{tabular}{|c|c|c|ccccccccccc|}
		\hline
			Dataset & Supervision & IoU threshold  $\rightarrow$   & 0.5           & 0.55          & 0.6           &  0.65          & 0.7           &   0.75          & 0.8           & 0.85         & 0.9          & 0.95         & Average \\ 
			\hline
			\multirow{12}*{1.2} & Full        & SSN     \cite{zhao2017temporal,shou2018autoloc}     & 41.3          & 38.8          & 35.9          & 32.9          & 30.4          & 27.0          & 22.2          & 18.2         & 13.2         & 6.1          & 26.6  \\ 
			\cline{2-14}
			~ & Weak        & UntrimmedNets \cite{wang2017untrimmednets} & 7.4           & 6.1           & 5.2           & 4.5           & 3.9           &  3.2           & 2.5           & 1.8          & 1.2          & 0.7          & 3.6   \\
			
			~ & Weak        & AutoLoc \cite{shou2018autoloc}     & 27.3 &  24.9 & 22.5 & 19.9 & 17.5 &  15.1 & 13.0 & 10.0 & 6.8 & 3.3 & 16.0 \\
			~ & Weak        & TSM \cite{yu2019temporal} & 28.3 & 26.0 & 23.6 & 21.2 & 18.9 & 17.0 & 14.0 & 11.1 & 7.5 &  3.5        &  17.1 \\
			
			~ & Weak        & WTALC \cite{paul2018w} & 37.0         & -           & -           & -           & 14.6          &  -           & -           & -          & -          & -          &  18.0 \\
			~ & Weak        & CleanNet \cite{Liu_2019_ICCV} & 37.1         & 33.4           & 29.9           & 26.7           & 23.4         &  20.3          & 17.2           & 13.9          & 9.2          & 5.0          &  21.6 \\

			%Weak        & \textbf{TTC-Loc (Ours)}     & \textbf{37.6} & \textbf{34.6} & \textbf{31.6} & \textbf{28.7} & \textbf{25.6} &  \textbf{22.6} & \textbf{19.6} & \textbf{15.3} & \textbf{10.9} & \textbf{4.9} & \textbf{23.1} \\
			%\hline
			
			~ & Semi (0)       & \textbf{TTC-Loc (Ours)}     & \textbf{39.4} & \textbf{36.1} & \textbf{33.1} & \textbf{30.0} & \textbf{26.6} &  \textbf{23.6} & \textbf{19.9} & \textbf{15.4} & \textbf{10.7} & \textbf{5.3} & \textbf{24.0} \\
			
			\cline{2-14}
			~ & Weak  + Count Labels       & 3C-Net \cite{narayan20193cnet} & 37.2          & -           & -           & -           & 23.7          &  -           & -           & -          & 9.2          & -          &  21.7 \\
			~ & Weak + Static Clips & Liu \textit{et al.}~\cite{Liu_2019_CVPR} & 36.8          & -           & -           & -           & -          &  22.0          & -           & -          & -          & 5.6         &  22.4 \\
			~ & Semi (1)       & \textbf{TTC-Loc (Ours)}     & \textbf{40.6} & \textbf{36.6} & \textbf{33.3} & \textbf{30.3} & \textbf{27.0} &  \textbf{23.6} & \textbf{20.6} & \textbf{16.2} & \textbf{11.3} & \textbf{5.3} & \textbf{24.5} \\
			%~ & Semi (2)       & \textbf{TTC-Loc (Ours)}     & \textbf{40.6} & \textbf{37.0} & \textbf{33.9} & \textbf{30.7} & \textbf{27.4} &  \textbf{23.9} & \textbf{20.4} & \textbf{16.5} & \textbf{11.8} & \textbf{5.6} & \textbf{24.8} \\
			%~ & Semi (3)       & \textbf{TTC-Loc (Ours)}     & \textbf{41.4} & \textbf{37.6} & \textbf{34.1} & \textbf{31.1} & \textbf{28.2} &  \textbf{24.7} & \textbf{21.4} & \textbf{16.6} & \textbf{11.2} & \textbf{5.7} & \textbf{25.2} \\
			\hline
			\multirow{10}*{1.3} & Full        & CDC \cite{cdc_zheng_cvpr17}     & 45.3        & -           & -           & -           & -          &  26.0      & -           & -          & -          & 0.2         &  23.8 \\ 
			\cline{2-14}
			%~ & Weak        & UntrimmedNets \cite{wang2017untrimmednets} & 7.4           & 6.1           & 5.2           & 4.5           & 3.9           &  3.2           & 2.5           & 1.8          & 1.2          & 0.7          & 3.6   \\
			
			%~ & Weak        & AutoLoc \cite{shou2018autoloc}     & 27.3 &  24.9 & 22.5 & 19.9 & 17.5 &  15.1 & 13.0 & 10.0 & 6.8 & 3.3 & 16.0 \\
			~ & Weak        & TSM \cite{yu2019temporal} & 30.3        & -           & -           & -           & -          &  19.0      & -           & -          & -          & 4.5         &  - \\
			
			~ & Weak        & STPN \cite{nguyen2018weakly} & 29.3        & -           & -           & -           & -          &  16.9       & -           & -          & -          & 2.6          &  - \\
			~ & Weak        & ST-GradCAM \cite{xu2018segregated} & 31.1         & -           & -          & -       & -    & 18.8     & -                & -          & -          & 4.7         &  - \\

			%Weak        & \textbf{TTC-Loc (Ours)}     & \textbf{37.6} & \textbf{34.6} & \textbf{31.6} & \textbf{28.7} & \textbf{25.6} &  \textbf{22.6} & \textbf{19.6} & \textbf{15.3} & \textbf{10.9} & \textbf{4.9} & \textbf{23.1} \\
			%\hline
			
			~ & Semi (0)       & \textbf{TTC-Loc (Ours)}     & \textbf{37.5} & \textbf{33.1} & \textbf{30.2} & \textbf{26.7} & \textbf{23.9} &  \textbf{20.6} & \textbf{17.4} & \textbf{14.1} & \textbf{9.9} & \textbf{4.5} & \textbf{21.8} \\
			
			\cline{2-14}
			~ & Weak  + MV       & Nguyen \textit{et al.} & 36.4          & -           & -           & -           & -          & 19.2          & -           & -          & 2.9          & -          &  - \\
			~ & Weak + Static Clips & Liu \textit{et al.}~\cite{Liu_2019_CVPR} & 34.0          & -           & -           & -           & -          &  20.9         & -           & -          & -          & 5.7         &  21.2 \\
			~ & Semi (1)       & \textbf{TTC-Loc (Ours)}     & \textbf{37.6} & \textbf{33.6} & \textbf{30.5} & \textbf{27.4} & \textbf{24.3} &  \textbf{21.5} & \textbf{18.0} & \textbf{14.4} & \textbf{9.9} & \textbf{4.7} & \textbf{22.2} \\
			%~ & Semi (2)       & \textbf{TTC-Loc (Ours)}     & \textbf{40.6} & \textbf{37.0} & \textbf{33.9} & \textbf{30.7} & \textbf{27.4} &  \textbf{23.9} & \textbf{20.4} & \textbf{16.5} & \textbf{11.8} & \textbf{5.6} & \textbf{24.8} \\
			%~ & Semi (3)       & \textbf{TTC-Loc (Ours)}     & \textbf{41.4} & \textbf{37.6} & \textbf{34.1} & \textbf{31.1} & \textbf{28.2} &  \textbf{24.7} & \textbf{21.4} & \textbf{16.6} & \textbf{11.2} & \textbf{5.7} & \textbf{25.2} \\
        
        \hline
        
		\end{tabular}
		
	\end{center}
	
\end{table*}

\subsection{Comparison with the State-of-the-art}
\textbf{THUMOS'14.}
Table \ref{table:res_th} summarizes the comparisons between our TTC-Loc model and state-of-the-art methods for temporal action localization on THUMOS'14 test set. \textbf{Under weakly-supervised setting, namely Semi(0)}, with UntrimmedNets features, our TTC-Loc method, achieves much better results than the state-of-the-art weakly-supervised methods; with I3D features, our method significantly outperforms the state-of-the-art under most IoU thresholds. For example, when IoU threshold is 0.5, our TTC-Loc model achieves $5.9\%$ absolute improvement (relatively $26\%$) over ST-GradCAM \cite{xu2018segregated}\footnote{Results under 0.6 and 0.7 are from E-mail communication with the authors.}. It is also encouraging that our TTC-Loc framework even outperforms the results reported in some recent fully-supervised temporal action localization methods, e.g., CDC \cite{cdc_zheng_cvpr17} by a large margin.

\textbf{Under Semi-supervised setting}, we compare our proposed TTC-Loc with the methods that are supervised with more supervision, e.g., Static Clips~\cite{Liu_2019_CVPR}, Count Labels \cite{narayan20193cnet} and MV \cite{nguyen2019weaklysupervised}, and fully-supervised methods. Our proposed TTC-Loc consistently enjoys a large performance boost with both kinds of features. When using features extracted by UntrimmedNets (sharing the same backbone with SSN), our TTC-Loc also significantly reduces the performance gap with SSN by being directly supervised by Semi-supervision without train-test discrepancy. Using I3D features, Nguyen \textit{et al.} \cite{nguyen2019weaklysupervised} achieved 28.2\% average mAP with the help of MV, which is on par with the performance of our TTC-Loc with only weak supervision. When only one sample of each class is fully annotated, our TTC-Loc is significantly improved by 3.8\% average mAP and even outperforms SSN    \cite{zhao2017temporal,shou2018autoloc} by a large margin.

\textbf{ActivityNet 1.2 and 1.3.}
Table~\ref{table:res_an} shows the results of our TTC-Loc framework and the compared state-of-the-art TAL methods on ActivityNet 1.2 and 1.3. \textbf{Under weakly-supervised setting, namely Semi(0)}, our TTC-Loc method again significantly outperforms state-of-the-art WTAL methods on both ActivityNet 1.2 and 1.3, which again verifies the effectiveness of our TTC-Loc framework. \textbf{Under semi-supervised setting,} our TTC-Loc is consistently improved with the help of only a few fully-annotated samples. Note that ActivityNet 1.2 and 1.3 are much larger than THUMOS'14, therefore one sample per class means that only about 2\% samples are annotated with temporal boundaries. When compared to other methods that uses extra supervision, our TTC-Loc significantly outperforms them with only one fully annotated sample per class. For example, when IoU threshold is 0.7, the proposed TTC-Loc outperforms 3C-Net~\cite{narayan20193cnet} by $3.3\%$, and the relative improvement is about $14\%$. In terms of the average mAP, our TTC-Loc model also improves the state-of-the-art \cite{Liu_2019_CVPR} by a large margin, $2.1\%$. It is also noteworthy that Our TTC-Loc with semi-supervision even achieves comparable results with fully-supervised methods. For example, on ActivityNet 1.3, our TTC-Loc is very close to CDC~\cite{cdc_zheng_cvpr17} in terms of average mAP and even outperforms CDC when IoU threshold is high.

%Note that in both of the comparisons on THUMOS'14 and ActivityNet 1.2, we compare our TTC-Loc methods with methods using the similar architectures.  ST-GradCAM~\cite{xu2018segregated} and Liu \textit{et al.}~\cite{Liu_2019_CVPR} use RNN and TemConv respectively to capture high-level temporal patterns. Therefore, we compare them with our TTC-Loc+TemConv. In conclusion, our TTC-Loc framework is compatible with different architectures and different backbone feature extractor (UNtrimmedNets or I3D), and achieves the new state-of-the-art on two benchmark datasets without other data augmentation and post-processing techniques.

%We observe that on both THUMOS'14 and ActivityNet 1.2, our TTC-Loc method does not achieve the best performance when IoU threshold is small, although already the second best. We argue that this is a trade-off between the performance when IoU threshold is large and the performance when IoU threshold is small. It is noticeable that those methods, i.e., ST-GradCAM and WTALC, have a significantly larger drop than our TTC-Loc model when IoU threshold increases.

\begin{table}[!t]
\tiny
\caption{Importance of Train-Test Consistency. Train/test time strategy indicates the strategy for localizing action segments. mAP (\%) under different IoU thresholds on THUMOS'14 test set is reported.}
	\label{table:abth}
	\begin{center}

	\begin{tabular}{|c|c|c|ccccc|}
	\hline
	 Train time strategy & Test time Strategy & IoU threshold $\rightarrow$     & 0.3          & 0.4           & 0.5         & 0.6         & 0.7         \\ \hline
	 None & Manually Set \cite{paul2018w} & Weak & 40.1          & 31.1          & 22.8          & 14.8             & 7.6            \\
	Predicted & Manually Set \cite{paul2018w} & Weak & 40.6           & 31.7          & 24.6           & 14.8          & 8.0         \\
	Predicted  & Predicted    & Weak    & \textbf{46.7}   & \textbf{37.6} & \textbf{28.9} & \textbf{17.7} & \textbf{10.0}   \\
	\hline
	 Manually Set \cite{paul2018w} & Manually Set \cite{paul2018w} & Semi (1)  & 41.3           & 32.0          & 24.9           & 15.8          & 8.1         \\
	 Predicted & Manually Set \cite{paul2018w} & Semi (1)  & 41.0           & 31.9          & 24.8           & 15.7         & 7.4         \\
	
	Predicted  & Predicted    & Semi (1)    & \textbf{50.9}  & \textbf{41.9} & \textbf{33.4} & \textbf{21.7} & \textbf{12.1}    \\
	\hline
	
	\end{tabular}
	
	\end{center}
	
\end{table}

\subsection{Analysis}

\subsubsection{Importance of Train-Test Consistency and Predicted Threshold.}
\label{sec:lat}
Table~\ref{table:abth} shows the results of the same TTC-Loc model trained on THUMOS'14 with different train/test action localization strategies. Manually Set means setting the threshold as~\cite{paul2018w}, which is the average of maximum score and minimum score of one action class in one video. Predicted indicates our default setting. None indicates that in ~\cite{paul2018w}, during training, action segments are not localized and only Top-K scores are selected from each class to be fused for video classification. With weak supervision, our TTC-Loc using predicted threshold for localizing action segments during both training and testing outperforms inconsistent variants by a large margin. More importantly, under semi-supervised setting, the margin between our TTC-Loc and inconsistent variances is larger than that of weakly-supervised setting, which verifies that train-test discrepancy of action localization prevents existing WTAL models to benefit from semi-supervision.

We also evaluate \cite{paul2018w} in a semi-supervised setting by creating a soft localization gate with the Manually Set threshold that it uses at test time. As shown in Table~\ref{table:abth}, the semi-supervised version (Manually Set/Manually Set) improves over the weakly-supervised version (None/Manually Set). However, the improvement is till much smaller than that of our TTC-Loc, which validates of the effectiveness of our framework on adapt semi-supervision.

%As shown in Table~\ref{table:abth}, the Predicted Threshold consistently outperforms manually set threshold, which indicates that learning to predict the threshold is a better way to localize action segments.

\begin{figure*}[t]
\centering
\includegraphics[width=0.99\textwidth]{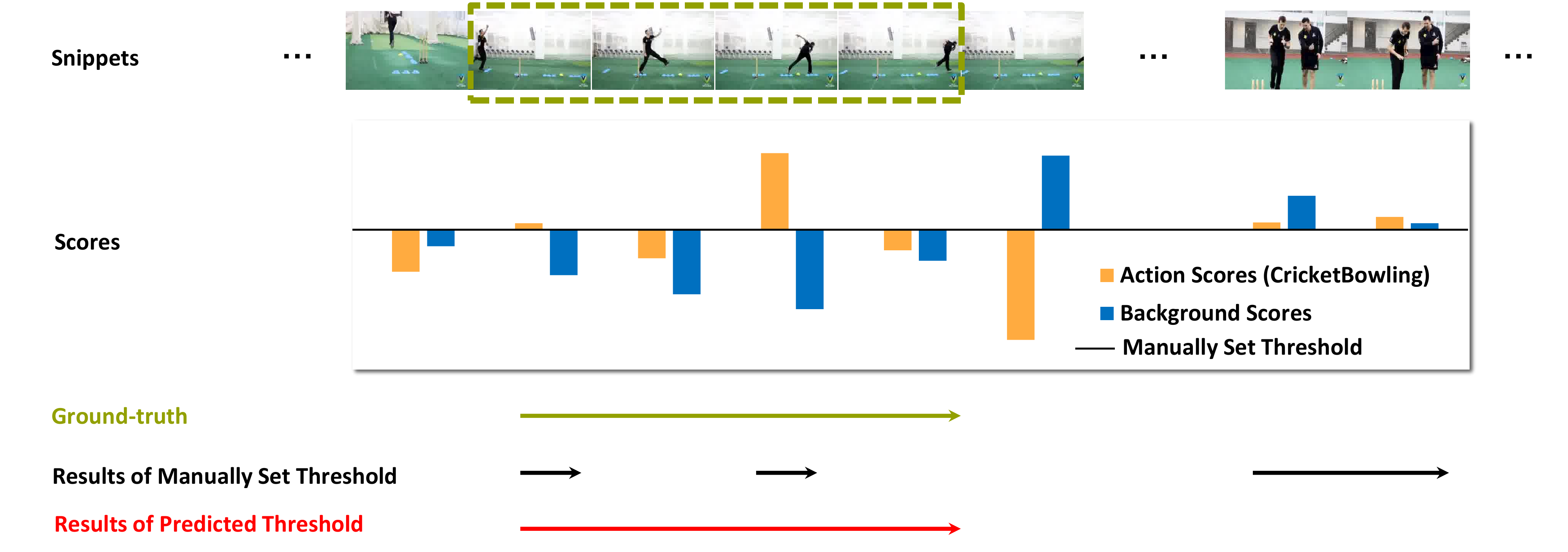}
\caption{Visualization of parts of the localization results on Video 1313 in the test set of THUMOS' 14. (Best viewed in color when zoomed in)
}
\label{fig:visual}
\end{figure*}

In Fig. \ref{fig:visual}, we visualize part of the localization results of Video 1313 in the test set of THUMOS'14. The ground truth segment does not have consistently high action scores for CricketBowling, the ground truth label of this segment. This is probably because this view point is relatively rare in the dataset. If we follow \cite{paul2018w} to set the threshold manually, only part of the segment is detected. However, using the Predicted Threshold, we can detect the whole segment. 

Our TTC-Loc model with the Predicted Threshold is also capable of reducing false alarms like the second segment, as is shown in Fig. \ref{fig:visual}. In this segment, the athlete keeps holding the cricket ball without bowling it, but it does look like bowling. Manually Set Threshold will falsely detect this segment. However, with our Predicted Adaptive Threshold, the background also has a relatively high score since the action is not clear. By using the Predicted Threshold, we can avoid this false alarm.

\subsubsection{Ablation Studies.}
\label{ab}
In this section, we explore the effects of the three loss terms in our formulation. We train a model with only video-level classification loss in different formulations. We select two representative formulations for comparison. In \cite{nguyen2018weakly}, a class-agnostic attention score is used to allocate scores from different snippets for video-level action classification. In \cite{paul2018w}, the idea of multiple instance learning is borrowed and for each action class, the highest $\frac{1}{8}$ scores are averaged to get the video level score for action classification. As Table~\ref{tab:loss} shows, when with only $L_{clas}$, our $L_{clas}$ achieves the best performance among different formulations. Note that with only our $L_{clas}$, the model can achieve comparable performance with STPN~\cite{nguyen2018weakly}, which leverages a sparse constraint to boost the model's localization ability. The superiority shown in this comparison strongly supports the effectiveness of our formulation and the importance of bridging the gap between training and testing in previous methods.

As is expected, under both weakly-supervised setting and semi-supervised setting, adding $L_{reg}$ significantly improves the localization performance because $L_{reg}$ regularizes the model to produce large margins between actions scores and background scores, which results in action proposals of high robustness. We do not conduct experiments without the video classification loss because this is the only supervision we get in WTAL.

We also explore different semi-supervised training strategy to optimize our model. Only training our TTC-Loc on fully annotated subset produces relatively low performance because there is only one sample in each class. Pre-training on the weakly annotated subset and fine-tuning on the fully-annotated set is better but still worse than our default strategy, namely, joint training. 

\begin{table}[t] 
\caption{Temporal localization average mAP (\%) under IoU threshold from 0.3 to 0.7 on THUMOS'14 test set of different variants. Semi indicates one video per class is fully annotated and the rest videos only have video action labels.}
\tiny
  \begin{minipage}[b]{0.5\textwidth} 
    %\tiny
    \centering
    \subcaption{Loss terms}
    \label{tab:loss}
    \begin{tabular}{|c|c|}
    	\hline
    	 Objectives  & Avg     \\ \hline
    	  $L_{clas}$~\cite{nguyen2018weakly} & 14.8  \\
    	 $L_{clas}$ \cite{paul2018w} & 17.0  \\
    	
    	$L_{clas}$ (Ours)       & 18.4     
    	\\
    	$L_{clas}$ and $L_{reg}$ (Ours, Weak)       & 28.2  
    	\\
    	$L_{clas}$ and $L_{loc}$ (Ours, Semi)       & 28.5 
    	\\
    	$L_{clas}$, $L_{reg}$ and $L_{loc}$ (Ours, Semi)       & 32.0   
    	\\
    	\hline

        \end{tabular} 
  \end{minipage}% 
  \begin{minipage}[b]{0.5\textwidth} 
    %\tiny
    \centering
    \subcaption{Semi-supervised training strategies}
    \label{tab:sem}
    \begin{tabular}{|c|c|}
    	\hline
    	 Training Strategy  & Avg     \\ \hline
    	  Fully Annotated Only & 22.9  \\
    	 Pre-train + Fine-tune & 28.4  \\
    	
    	\textbf{Joint-Train (Default)}      & 32.0    
    	\\
    	
    	\hline

        \end{tabular}
  \end{minipage} 
  
  \begin{minipage}[b]{0.5\textwidth} 
    %\tiny
    \centering
    \subcaption{Gating Function}
    \label{gate}
    \begin{tabular}{|c|c|}
	\hline
	 Gating Function  & Avg     \\ \hline
	  %ReLU & 0 \\
	Binarization & 13.5
	\\
	Softsign & 27.4 \\
	\textbf{Sigmoid (Default)}       & 28.2   
	\\
	\hline

    \end{tabular}
  \end{minipage}% 
  \begin{minipage}[b]{0.5\textwidth} 
    %\tiny
    \centering
    \subcaption{Formulation of $L_{reg}$}
    \label{lcont}
    \begin{tabular}{|c|c|}
	\hline
	 Formulation  & Avg     \\ \hline
	  L1 Norm & 20.4  \\
	 L2 Norm & 21.1  \\
	
	Cosine Similarity       & 19.6     
	\\
	\textbf{Inner Product (Default)}       & 28.2 
	\\
	\hline

    \end{tabular}
  \end{minipage} 
  \vspace{-0.4cm}
\end{table}

\subsubsection{Detailed Analysis on the Gating Function and the Form of $L_{reg}$.}
\label{gate_form}
We explore different gating functions to obtain the gate $g^{(i)}$. In deep hashing/binary encoding, besides Sigmoid function, other relaxation functions are also proposed for learning to predict binary codes~\cite{erin2015deep,duan2018graphbit,karaman2019unsupervised}. We also test binarization and softsign\footnote{$f(x)=\frac{x/(1+|x|)+1}{2}$}. For binarization, we binarize the input into either 0 or 1 in the forward call, but in the backward call, we use the gradient of identity transformation as the surrogate of the gradient of the step function. As Table~\ref{gate} shows, Binarization does not work well. Binarization suffers from oscillation around a local minimum and is hard to converge. Softsign and sigmoid are both continuous and easier to optimize but sigmoid achieves the better result, which justifies the choice of our gating function.

%\label{loss_form}
We explore different formulations of $L_{reg}$. In our default setting, we penalize the inner product of ground truth action scores and background scores. This formulation guarantees a large margin between these two scores. There are other distance metrics which can be used to enforce a large margin. We try replacing inner product with L1 Norm, L2 Norm, and Cosine Similarity. As Table~\ref{lcont} shows, our default setting achieves the best localization results on THUMOS'14. We assume this is because cosine similarity brings extra constraint on the scores from different snippets. L1 and L2 norms will not enforce the margin to be large enough. For example, when ground truth action score of a snippet is 0.2 and background score is $-4$, L1 and L2 norms will not further maximize the margin because the distance is already larger than 1. But our inner product formulation will further push the distance to be less than -1.

\section{Conclusion}
In this paper, we propose a Train-Test Consistent framework, TTC-Loc, which leverages semi-supervision to boost performance of weakly supervised temporal action localization models. We recognize the large gap between weakly-supervised and fully-supervised models, and first propose to utilize full annotations of a few examples to bridge the gap. We identify the train-test discrepancy of action localization prevents WTAL models to be explicitly supervised by temporal annotations and propose to predict thresholds for consistent action localization behavior at training and testing. Under weak supervision, our TTC-Loc significantly outperforms the state-of-the-art performance on THUMOS'14, ActivityNet 1.2 and 1.3; with semi-supervision, our TTC-Loc further boosts the performance and achieve comparable performance with some fully-supervised methods.
\bibliographystyle{splncs04}
\bibliography{egbib}
\end{document}